\documentclass[runningheads]{llncs}
\usepackage{graphicx}

\usepackage{tikz}
\usepackage{comment} 
\usepackage{amsmath,amssymb} 
\usepackage{color}
\usepackage{multirow}
\usepackage{amsfonts}
\usepackage{booktabs}
\usepackage{siunitx,booktabs,caption}

\usepackage[width=122mm,left=12mm,paperwidth=146mm,height=193mm,top=12mm,paperheight=217mm]{geometry}

\begin{document}
\pagestyle{headings}
\mainmatter
\def\ECCVSubNumber{3526}  


\title{RAFT: Recurrent All-Pairs Field Transforms for Optical Flow}
\titlerunning{RAFT: Recurrent All-Pairs Field Transforms}

\author{Zachary Teed \and Jia Deng}
\authorrunning{Z. Teed and J. Deng}
\institute{Princeton University \\ 
\email{\{zteed,jiadeng\}@cs.princeton.edu}}

\maketitle

\begin{abstract}
We introduce Recurrent All-Pairs Field Transforms (RAFT), a new deep network architecture for optical flow. RAFT extracts per-pixel features, builds multi-scale 4D correlation volumes for all pairs of pixels, and iteratively updates a flow field through a recurrent unit that performs lookups on the correlation volumes. RAFT achieves state-of-the-art performance. On KITTI, RAFT achieves an F1-all error of 5.10\%, a 16\% error reduction from the best published result (6.10\%). On Sintel (final pass), RAFT obtains an end-point-error of 2.855 pixels, a 30\% error reduction from the best published result (4.098 pixels). In addition, RAFT has strong cross-dataset generalization as well as high efficiency in inference time, training speed, and parameter count. Code is available at \url{https://github.com/princeton-vl/RAFT}.
\end{abstract}

\section{Introduction}
Optical flow is the task of estimating per-pixel motion between video frames. It is a long-standing vision problem that remains unsolved. The best systems are limited by difficulties including fast-moving objects, occlusions, motion blur, and textureless surfaces. 

Optical flow has traditionally been approached as a hand-crafted optimization problem over the space of dense displacement fields between a pair of images \cite{horn1981determining,tvl1,fullflow}. Generally, the optimization objective defines a trade-off between a \emph{data} term which encourages the alignment of visually similar image regions and a \emph{regularization} term which imposes priors on the plausibility of motion. Such an approach has achieved considerable success, but further progress has appeared challenging, due to the difficulties in hand-designing an optimization objective that is robust to a variety of corner cases. 

Recently, deep learning has been shown as a promising alternative to traditional methods. Deep learning can side-step formulating an optimization problem and train a network to directly predict flow. Current deep learning methods~\cite{ilg2017flownet,pwcnet,liteflownet,vcn,flowelements} have achieved performance comparable to the best traditional methods while being significantly faster at inference time. A key question for further research is designing effective architectures that perform better, train more easily and generalize well to novel scenes. 

\begin{figure}[t]
	\begin{center}
		\includegraphics[width=\columnwidth]{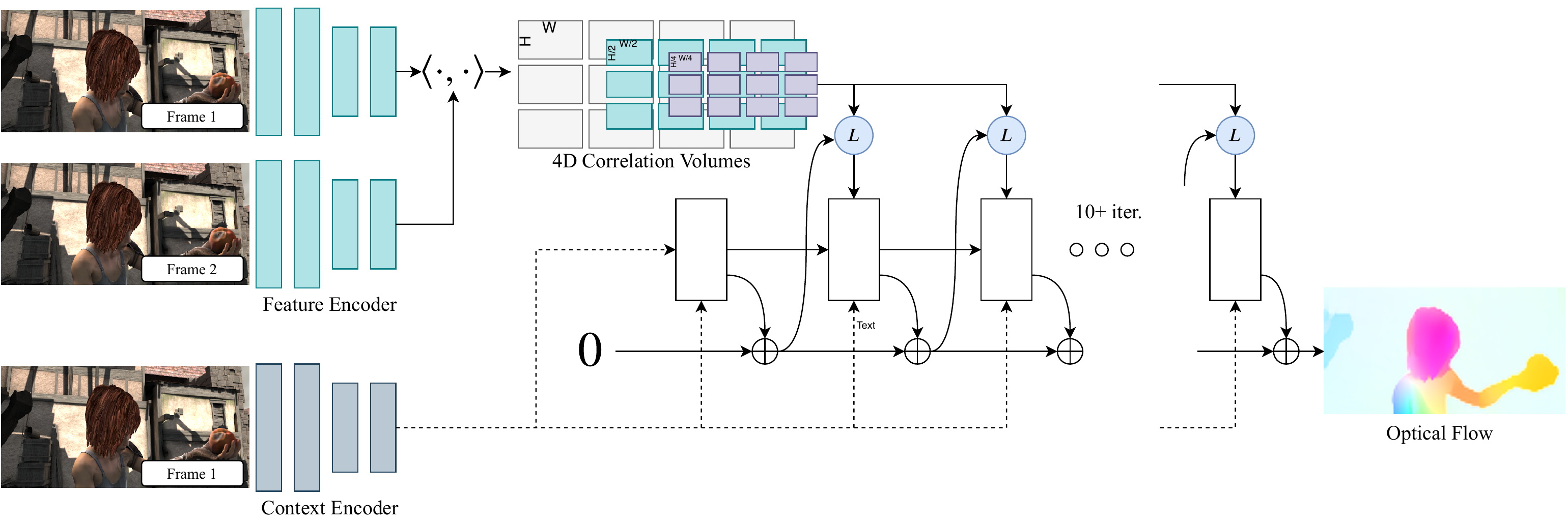}
	\end{center}
	\caption{RAFT consists of 3 main components: (1) A feature encoder that extracts per-pixel features from both input images, along with a context encoder that extracts features from only $I_1$. (2) A correlation layer which constructs a 4D $W\times H \times W \times H$ correlation volume by taking the inner product of all pairs of feature vectors. The last 2-dimensions of the 4D volume are pooled at multiple scales to construct a set of multi-scale volumes. (3) An \emph{update operator} which recurrently updates optical flow by using the current estimate to look up values from the set of correlation volumes.}
	\label{fig:Method}
\end{figure}

We introduce Recurrent All-Pairs Field Transforms (RAFT), a new deep network architecture for optical flow. RAFT enjoys the following strengths: 
\begin{itemize}
    \item \emph{State-of-the-art accuracy}: On KITTI~\cite{kitti}, RAFT achieves an F1-all error of 5.10\%, a 16\% error reduction from the best published result (6.10\%). On Sintel~\cite{sintel} (final pass), RAFT obtains an end-point-error of 2.855 pixels, a 30\% error reduction from the best published result (4.098 pixels). 
    \item \emph{Strong generalization}: When trained only on synthetic data, RAFT achieves an end-point-error of 5.04 pixels on KITTI~\cite{kitti}, a 40\% error reduction from the best prior deep network trained on the same data (8.36 pixels). 
    \item \emph{High efficiency}: RAFT processes $1088\times436$ videos at 10 frames per second on a 1080Ti GPU. It trains with 10X fewer iterations than other architectures. A smaller version of RAFT with 1/5 of the parameters runs at 20 frames per second while still outperforming all prior methods on Sintel. 
\end{itemize} 

RAFT consists of three main components: (1) a feature encoder that extracts a feature vector for each pixel; (2) a correlation layer that produces a 4D correlation volume for all pairs of pixels, with subsequent pooling to produce lower resolution volumes; (3) a recurrent GRU-based \emph{update operator} that retrieves values from the correlation volumes and iteratively updates a flow field initialized at zero. Fig.~\ref{fig:Method} illustrates the design of RAFT. 

The RAFT architecture is motivated by traditional optimization-based approaches. The feature encoder extracts per-pixel features. The correlation layer computes visual similarity between pixels. The update operator mimics the steps of an iterative optimization algorithm. But unlike traditional approaches, features and motion priors are not handcrafted but learned---learned by the feature encoder and the update operator respectively. 

The design of RAFT draws inspiration from many existing works but is substantially novel. First, RAFT maintains and updates a single fixed flow field at high resolution. This is different from the prevailing coarse-to-fine design in prior work~\cite{pwcnet,vcn,liteflownet,liteflownet2,hd3}, where flow is first estimated at low resolution and upsampled and refined at high resolution. By operating on a single high-resolution flow field, RAFT overcomes several limitations of a coarse-to-fine cascade: the difficulty of recovering from errors at coarse resolutions, the tendency to miss small fast-moving objects, and the many training iterations (often over 1M) typically required for training a multi-stage cascade. 

Second, the update operator of RAFT is recurrent and lightweight. Many recent works~\cite{irr,pwcnet,vcn,liteflownet,ilg2017flownet} have included some form of iterative refinement, but do not tie the weights across iterations~\cite{pwcnet,vcn,liteflownet} and are therefore limited to a fixed number of iterations. To our knowledge, IRR~\cite{irr} is the only  deep learning approach~\cite{irr} that is recurrent. It uses FlowNetS~\cite{flownet} or PWC-Net~\cite{pwcnet} as its recurrent unit. When using FlowNetS, it is limited by the size of the network (38M parameters) and is only applied up to 5 iterations. When using PWC-Net, iterations are limited by the number of pyramid levels. In contrast, our update operator has only 2.7M parameters and can be applied 100+ times during inference without divergence. 

Third, the update operator has a novel design, which consists of a convolutional GRU that performs lookups on 4D multi-scale correlation volumes; in contrast, refinement modules in prior work typically use only plain convolution or correlation layers. 

We conduct experiments on Sintel\cite{sintel} and KITTI\cite{kitti}. Results show that RAFT achieves state-of-the-art performance on both datasets. In addition, we validate various design choices of RAFT through extensive ablation studies.

\section{Related Work}

\noindent\textbf{Optical Flow as Energy Minimization}
Optical flow has traditionally been treated as an energy minimization problem which imposes a tradeoff between a \emph{data} term and a \emph{regularization} term. Horn and Schnuck \cite{horn1981determining} formulated optical flow as a continuous optimization problem using a variational framework, and were able to estimate a dense flow field by performing gradient steps. Black and Anandan\cite{black1993framework} addressed problems with oversmoothing and noise sensitivity by introducing a robust estimation framework.  TV-L1 \cite{tvl1} replaced the quadratic penalties with an L1 data term and total variation regularization, which allowed for motion discontinuities and was better equipped to handle outliers. Improvements have been made by defining better matching costs~\cite{deepflow,largedisplacement} and regularization terms~\cite{totalflow}.

Such continuous formulations maintain a single estimate of optical flow which is refined at each iteration.  To ensure a smooth objective function, a first order Taylor approximation is used to model the data term. As a result, they only work well for small displacements. To handle large displacements, the coarse-to-fine strategy is used, where an image pyramid is used to estimate large displacements at low resolution, then small displacements refined at high resolution. But this coarse-to-fine strategy may miss small fast-moving objects and have difficulty recovering from early mistakes. Like continuous methods, we maintain a single estimate of optical flow which is refined with each iteration. However, since we build correlation volumes for all pairs at both high resolution and low resolution, each local update uses information about both small and large displacements. In addition, instead of using a subpixel Taylor approximation of the data term, our update operator learns to propose the descent direction. 

More recently, optical flow has also been approached as a discrete optimization problem \cite{discreteflow,fullflow,dcflow} using a global objective. One challenge of this approach is the massive size of the search space, as each pixel can be reasonably paired with thousands of points in the other frame. Menez et al\cite{discreteflow} pruned the search space using feature descriptors and approximated the global MAP estimate using message passing. Chen et al. \cite{fullflow} showed that by using the distance transform, solving the global optimization problem over the full space of flow fields is tractable.  DCFlow \cite{dcflow} showed further improvements by using a neural network as a feature descriptor, and constructed a 4D cost volume over all pairs of features. The 4D cost volume was then processed using the Semi-Global Matching (SGM) algorithm~\cite{sgm}. Like DCFlow, we also constructed 4D cost volumes over learned features. However, instead of processing the cost volumes using SGM, we use a neural network to estimate flow. Our approach is end-to-end differentiable, meaning the feature encoder can be trained with the rest of the network to directly minimize the error of the final flow estimate. In contrast, DCFlow requires their network to be trained using an embedding loss between pixels; it cannot be trained directly on optical flow because their cost volume processing is not differentiable. 

\smallskip\noindent \textbf{Direct Flow Prediction} Neural networks have been trained to directly predict optical flow between a pair of frames, side-stepping the optimization problem completely. Coarse-to-fine processing has emerged as a popular ingredient in many recent works~\cite{pwcnet,hd3,liteflownet,liteflownet2,irr,vcn,flowelements,scopeflow,maskflownet}. In contrast, our method maintains and updates a single high-resolution flow field.

\smallskip\noindent\textbf{Iterative Refinement for Optical Flow}
Many recent works have used iterative refinement to improve results on optical flow~\cite{ilg2017flownet,spynet,pwcnet,liteflownet,vcn} and related tasks~\cite{iresnet,deeptam,deepv2d,li2018recurrent}. Ilg et al. \cite{ilg2017flownet} applied iterative refinement to optical flow by stacking multiple FlowNetS and FlowNetC modules in series. SpyNet\cite{spynet}, PWC-Net\cite{pwcnet}, LiteFlowNet\cite{liteflownet}, and VCN \cite{vcn} apply iterative refinement using coarse-to-fine pyramids. The main difference of these approaches from ours is that they do not share weights between iterations. 

More closely related to our approach is IRR\cite{irr}, which builds off of the FlownetS and PWC-Net architecture but shares weights between refinement networks.  When using FlowNetS, it is limited by the size of the network (38M parameters) and is only applied up to 5 iterations. When using PWC-Net, iterations are limited by the number of pyramid levels. In contrast, we use a much simpler refinement module (2.7M parameters) which can be applied for 100+ iterations during inference without divergence. Our method also shares similarites with Devon \cite{devon}, namely the construction of the cost volume without warping and fixed resolution updates. However, Devon does not have any recurrent unit. It also differs from ours regarding large displacements. Devon handles large displacements using a dilated cost volume while our approach pools the correlation volume at multiple resolutions.

Our method also has ties to TrellisNet \cite{trellis} and Deep Equilibrium Models (DEQ) \cite{deq}. Trellis net uses depth tied weights over a large number of layers, DEQ simulates an infinite number of layers by solving for the fixed point directly. TrellisNet and DEQ were designed for sequence modeling tasks, but we adopt the core idea of using a large number of weight-tied units. Our update operator uses a modified GRU block\cite{cho2014properties}, which is similar to the LSTM block used in TrellisNet. We found that this structure allows our update operator to more easily converge to a fixed flow field.

\smallskip\noindent\textbf{Learning to Optimize}
Many problems in vision can be formulated as an optimization problem. This has motivated several works to embed optimization problems into network architectures \cite{optnet,diffcvx,banet,invcomp,deepv2d}. These works typically use a network to predict the inputs or parameters of the optimization problem, and then train the network weights by backpropogating the gradient through the solver, either implicitly\cite{optnet,diffcvx} or unrolling each step \cite{invcomp,banet}. However, this technique is limited to problems with an objective that can be easily defined.

Another approach is to learn iterative updates directly from data \cite{adler2017solving,adler2018learned}. These approaches are motivated by the fact that first order optimizers such as Primal Dual Hybrid Gradient (PDHG)\cite{pdhg} can be expressed as a sequence of iterative update steps. Instead of using an optimizer directly, Adler et al. \cite{adler2017solving} proposed building a network which mimics the updates of a first order algorithm. This approach has been applied to inverse problems such as image denoising \cite{variationalnetworks}, tomographic reconstruction \cite{adler2018learned}, and novel view synthesis\cite{deepview}. TVNet \cite{tvnet} implemented the TV-L1 algorithm as a computation graph, which enabled the training the TV-L1 parameters. However, TVNet operates directly based on intensity gradients instead of learned features, which limits the achievable accuracy on challenging datasets such as Sintel. 

Our approach can be viewed as learning to optimize: our network uses a large number of update blocks to emulate the steps of a first-order optimization algorithm. However, unlike prior work, we never explicitly define a gradient with respect to some optimization objective. Instead, our network retrieves features from correlation volumes to propose the descent direction.

\section{Approach}

Given a pair of consecutive RGB images, $I_1$, $I_2$, we estimate a dense displacement field $(\mathbf{f}^1, \mathbf{f}^2)$ which maps each pixel $(u, v)$ in $I_2$ to its corresponding coordinates $(u', v') = (u + f^1(u), v + f^2(v))$ in $I_2$. An overview of our approach is given in Figure \ref{fig:Method}. Our method can be distilled down to three stages: (1) feature extraction, (2) computing visual similarity, and (3) iterative updates, where all stages are differentiable and composed into an end-to-end trainable architecture. 

\subsection{Feature Extraction}
\label{sec:feature_extraction}
Features are extracted from the input images using a convolutional network. The feature encoder network is applied to both $I_1$ and $I_2$ and maps the input images to dense feature maps at a lower resolution. Our encoder, $g_\theta$ outputs features at 1/8 resolution $g_\theta: \mathbb{R}^{H \times W \times 3} \mapsto \mathbb{R}^{H/8 \times W/8 \times D}$ where we set $D=256$. The feature encoder consists of 6 residual blocks, 2 at 1/2 resolution, 2 at 1/4 resolution, and 2 at 1/8 resolution (more details in the supplemental material).

We additionally use a context network. The context network extracts features only from the first input image $I_1$. The architecture of the context network, $h_\theta$ is identical to the feature extraction network. Together, the feature network $g_\theta$ and the context network $h_\theta$ form the first stage of our approach, which only need to be performed once.

\subsection{Computing Visual Similarity}
\label{sec:similarity}

We compute visual similarity by constructing a full correlation volume between all pairs. Given image features $g_\theta(I_1) \in \mathbb{R}^{H \times W \times D}$ and $g_\theta(I_2) \in \mathbb{R}^{H \times W \times D}$, the correlation volume is formed by taking the dot product between all pairs of feature vectors. The correlation volume, $\mathbf{C}$, can be efficiently computed as a single matrix multiplication.
\begin{align}
    \mathbf{C}(g_\theta(I_1), g_\theta(I_2)) \in \mathbb{R}^{H \times W \times H \times W}, \qquad
    C_{ijkl} = \sum_h g_\theta(I_1)_{ijh} \cdot g_\theta(I_2)_{klh}
\end{align}
\begin{figure}[t]
    \centering
	\includegraphics[width=.9\columnwidth]{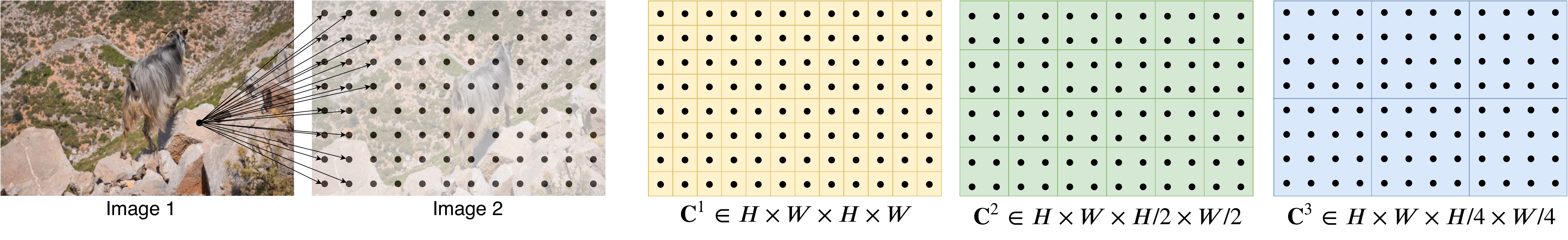}
	\caption{Building correlation volumes. Here we depict 2D slices of a full 4D volume. For a feature vector in $I_1$, we take take the inner product with all pairs in $I_2$, generating a 4D $W\times H \times W \times H$ volume (each pixel in $I_2$ produces a 2D response map). The volume is pooled using average pooling with kernel sizes $\{1, 2, 4, 8\}$.}
	\label{fig:CorrPyramid}
\end{figure}

\noindent \textbf{Correlation Pyramid:} We construct a 4-layer pyramid $\{\mathbf{C}^1, \mathbf{C}^2, \mathbf{C}^3, \mathbf{C}^4 \}$ by pooling the last two dimensions of the correlation volume with kernel sizes 1, 2, 4, and 8 and equivalent stride (Figure \ref{fig:CorrPyramid}). Thus, volume $\mathbf{C}^k$ has dimensions $H \times W \times H/2^k \times W/2^k$. The set of volumes gives information about both large and small displacements;  however, by maintaining the first 2 dimensions (the $I_1$ dimensions) we maintain high resolution information, allowing our method to recover the motions of small fast-moving objects.

\smallskip \noindent \textbf{Correlation Lookup: } We define a lookup operator $L_\mathbf{C}$ which generates a feature map by indexing from the correlation pyramid. Given a current estimate of optical flow $(\mathbf{f}^1, \mathbf{f}^2)$, we map each pixel $\mathbf{x}=(u, v)$ in $I_1$ to its estimated correspondence in $I_2$: $\mathbf{x}' = (u + f^1(u), v + f^2(v))$. We then define a local grid around $\mathbf{x}'$
\begin{equation}
    \mathcal{N}(\mathbf{x'})_r = \{\mathbf{x'} + \mathbf{dx} \ \vert \ \mathbf{dx} \in \mathbb{Z}^2, ||\mathbf{dx}||_1 \leq r \}
\end{equation}
as the set of integer offsets which are within a radius of $r$ units of $\mathbf{x'}$ using the L1 distance. We use the local neighborhood $\mathcal{N}(\mathbf{x'})_r$ to index from the correlation volume. Since $\mathcal{N}(\mathbf{x'})_r$ is a grid of real numbers, we use bilinear sampling. 

We perform lookups on all levels of the pyramid, such that the correlation volume at level $k$, $\mathbf{C}^k$, is indexed using the grid $\mathcal{N}(\mathbf{x'} / 2^k)_r$. A constant radius across levels means larger context at lower levels: for the lowest level, $k=4$ using a radius of 4 corresponds to a range of 256 pixels at the original resolution. The values from each level are then concatenated into a single feature map.


\smallskip\noindent \textbf{Efficient Computation for High Resolution Images: } The all pairs correlation scales $O(N^2)$ where $N$ is the number of pixels, but only needs to be computed once and is constant in the number of iterations $M$. However, there exists an equivalent implementation of our approach which scales $O(NM)$ exploiting the linearity of the inner product and  average pooling. Consider the cost volume at level $m$, $\mathbf{C}^m_{ijkl}$, and feature maps $g^{(1)} = g_\theta(I_1)$, $g^{(2)} = g_\theta(I_2)$:
\begin{equation*}
        \mathbf{C}^m_{ijkl} = \frac{1}{2^{2m}} \sum_p^{2^m} \sum_q^{2^m} \langle g^{(1)}_{i,j}, g^{(2)}_{2^mk+p,2^ml+q} \rangle 
        = \langle g^{(1)}_{i,j}, \frac{1}{2^{2m}} (\sum_p^{2^m} \sum_q^{2^m} g^{(2)}_{2^mk+p,2^ml+q}) \rangle
\end{equation*}
which is the average over the correlation response in the $2^m \times 2^m$ grid. This means that the value at $\mathbf{C}^m_{ijkl}$ can be computed as the inner product between the feature vector $g_\theta(I_1)_{ij}$ and $g_\theta(I_2)$ pooled with kernel size $2^m \times 2^m$.

In this alternative implementation, we do not precompute the correlations, but instead precompute the pooled image feature maps. In each iteration, we compute each correlation value on demand---only when it is looked up. This gives a complexity of $O(NM)$. 

We found empirically that precomputing all pairs is easy to implement and not a bottleneck, due to highly optimized matrix routines on GPUs---even for 1088x1920 videos it takes only 17\% of total inference time. Note that we can always switch to the alternative implementation should it become a bottleneck.

\subsection{Iterative Updates}
\label{sec:updates}

Our update operator estimates a sequence of flow estimates $\{\mathbf{f}_1, ..., \mathbf{f}_N \}$ from an initial starting point $\mathbf{f}_0 = \mathbf{0}$. With each iteration, it produces an update direction $\Delta \mathbf{f}$ which is applied to the current estimate: $\mathbf{f}_{k+1} = \Delta \mathbf{f} + \mathbf{f}_{k+1}$.

The update operator takes flow, correlation, and a latent hidden state as input, and outputs the update $\Delta \mathbf{f}$ and an updated hidden state. The architecture of our update operator is designed to mimic the steps of an optimization algorithm. As such, we used tied weights across depth and use bounded activations to encourage convergence to a fixed point. The update operator is trained to perform updates such that the sequence converges to a fixed point $\mathbf{f}_k \rightarrow \mathbf{f}^*$.

\smallskip \noindent \textbf{Initialization: } By default, we initialize the flow field to 0 everywhere, but our iterative approach gives us the flexibility to experiment with alternatives. When applied to video, we test \emph{warm-start} initialization, where optical flow from the previous pair of frames is forward projected to the next pair of frames with occlusion gaps filled in using nearest neighbor interpolation.

\smallskip \noindent \textbf{Inputs: } Given the current flow estimate $\mathbf{f}^k$, we use it to retrieve correlation features from the correlation pyramid as described in Sec.~\ref{sec:similarity}. The correlation features are then processed by 2 convolutional layers. Additionally, we apply 2 convolutional layers to the flow estimate itself to generate flow features. Finally, we directly inject the input from the context network. The input feature map is then taken as the concatenation of the correlation, flow, and context features. 

\smallskip \noindent \textbf{Update: } A core component of the update operator is a gated activation unit based on the GRU cell, with fully connected layers replaced with convolutions:
\begin{align}
&z_t = \sigma(\text{Conv}_{\text{3x3}}([h_{t-1}, x_t], W_z)) \\
&r_t = \sigma(\text{Conv}_{\text{3x3}}([h_{t-1}, x_t], W_r)) \\
&\tilde{h_t} = \tanh(\text{Conv}_{\text{3x3}}([r_t \odot h_{t-1}, x_t], W_h)) \\
& h_{t} = (1 - z_t) \odot h_{t-1} + z_t \odot \tilde{h_t}
\end{align}
where $x_t$ is the concatenation of flow, correlation, and context features previously defined. We also experiment with a separable ConvGRU unit, where we replace the $3 \times 3$ convolution with two GRUs: one with a $1\times5$ convolution and one with a $5\times1$ convolution to increase the receptive field without significantly increasing the size of the model.

\smallskip \noindent \textbf{Flow Prediction: } The hidden state outputted by the GRU is passed through two convolutional layers to predict the flow update $\Delta \mathbf{f}$. The output flow is at 1/8 resolution of the input image. During training and evaluation, we upsample the predicted flow fields to match the resolution of the ground truth.

\smallskip \noindent \textbf{Upsampling: } The network outputs optical flow at 1/8 resolution. We upsample the optical flow to full resolution by taking the full resolution flow at each pixel to be the convex combination of a 3x3 grid of its coarse resolution neighbors. We use two convolutional layers to predict a $H/8\times W/8\times(8\times8\times9)$ mask and perform softmax over the weights of the 9 neighbors. The final high resolution flow field is found by using the mask to take a weighted combination over the neighborhood, then permuting and reshaping to a $H \times W \times 2$ dimensional flow field. This layer can be directly implemented in PyTorch using the \texttt{unfold} function.

\subsection{Supervision}
We supervised our network on the $l_1$ distance between the predicted and ground truth flow over the full sequence of predictions, $\{\mathbf{f}_1, ..., \mathbf{f}_N \}$, with exponentially increasing weights. Given ground truth flow $\mathbf{f}_{gt}$, the loss is defined as
\begin{equation}
    \mathcal{L} = \sum_{i=1}^{N} \gamma^{N-i} ||\mathbf{f}_{gt} - \mathbf{f}_i||_1
\end{equation}
where we set $\gamma=0.8$ in our experiments.

\section{Experiments}
We evaluate RAFT on Sintel\cite{sintel} and KITTI\cite{kitti}. Following previous works, we pretrain our network on FlyingChairs\cite{flownet} and FlyingThings\cite{largedataset}, followed by dataset specific finetuning. Our method achieves state-of-the-art performance on both Sintel (both clean and final passes) and KITTI. Additionally, we test our method on 1080p video from the DAVIS dataset\cite{davis} to demonstrate that our method scales to  videos of very high resolutions. 

\smallskip \noindent \textbf{Implementation Details: }
RAFT is implemented in PyTorch\cite{pytorch}. All modules are initialized from scratch with random weights. During training, we use the AdamW\cite{adamw} optimizer and clip gradients to the range $[-1, 1]$. Unless otherwise noted, we evaluate after 32 flow updates on Sintel and 24 on KITTI. For every update, $\Delta \mathbf{f} + \mathbf{f}_{k}$, we only backpropgate the gradient through the $\Delta \mathbf{f}$ branch, and zero the gradient through the $\mathbf{f}_{k}$ branch as suggested by \cite{flowelements}.

\smallskip \noindent \textbf{Training Schedule: }
We train RAFT using two 2080Ti GPUs. We pretrain on FlyingThings for 100k iterations with a batch size of 12, then train for 100k iterations on FlyingThings3D with a batch size of 6. We finetune on Sintel for another 100k by combining data from Sintel\cite{sintel}, KITTI-2015 \cite{menze2015object}, and HD1K\cite{hd1k} similar to MaskFlowNet \cite{maskflownet} and PWC-Net+ \cite{pwcnet+}. Finally, we finetune on KITTI-2015 for an additionally 50k iterations using the weights from the model finetuned on Sintel. Details on training and data augmentation are provided in the supplemental material. For comparison with prior work, we also include results from our model when finetuning only on Sintel and only on KITTI.

\begin{figure}[t]
    \centering
	\includegraphics[width=.9\columnwidth]{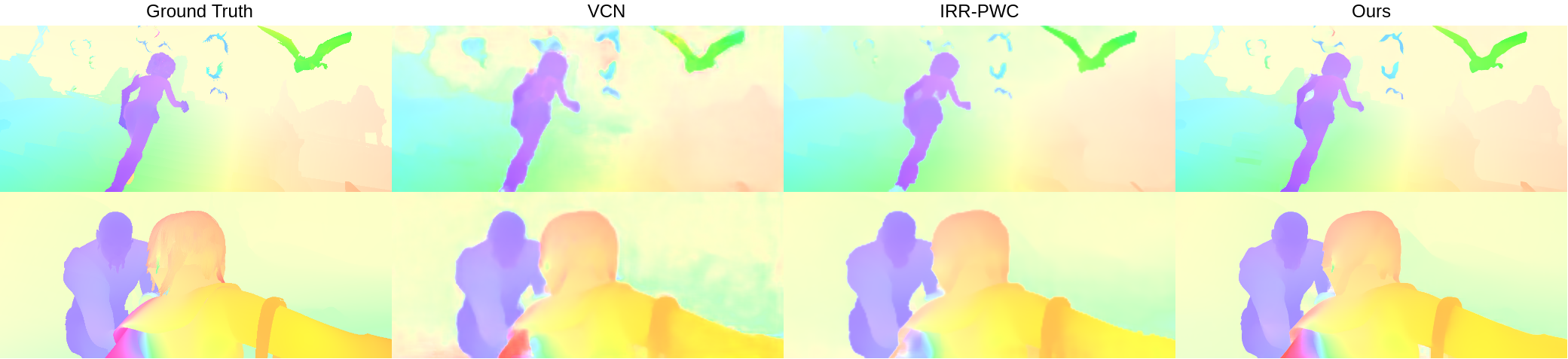}
	\caption{Flow predictions on the Sintel test set. }
	\label{fig:Sintel}
\end{figure}

\subsection{Sintel}
We train our model using the FlyingChairs$\rightarrow$FlyingThings schedule and then evaluate on the Sintel dataset using the \emph{train} split for validation. Results are shown in Table \ref{table:Results} and Figure \ref{fig:Sintel}, and we split results based on the data used for training. C + T means that the models are trained on FlyingChairs(C) and FlyingThings(T), while +ft indicates the model is finetuned on Sintel data. Like PWC-Net+\cite{pwcnet+} and MaskFlowNet \cite{maskflownet} we include data from KITTI and HD1K when finetuning. We train 3 times with different seeds, and report results using the model with the median accuracy on the clean pass of Sintel (train).

\begin{figure}[t]
    \centering
	\includegraphics[width=\columnwidth]{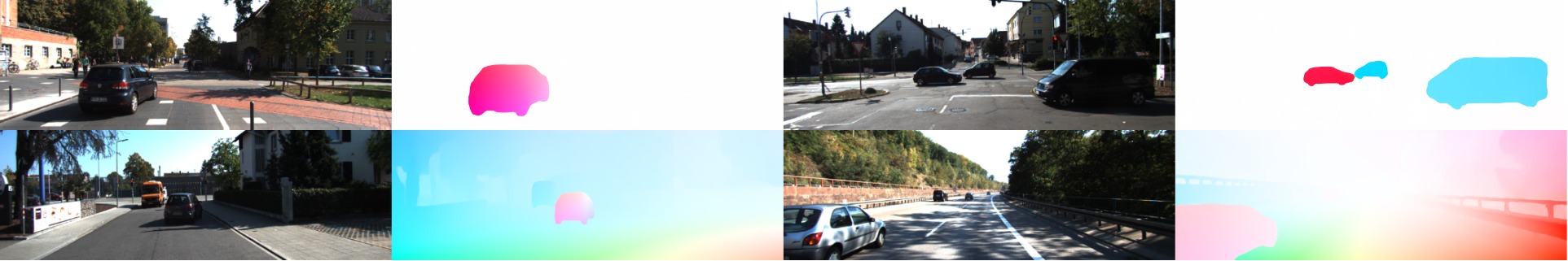}
	\caption{Flow predictions on the KITTI test set.}
	\label{fig:KITTI}
\end{figure}

\setlength\tabcolsep{.7em}
\begin{table}[h]
\centering
\resizebox{\textwidth}{!}{
\begin{tabular}{clccccccc}
\toprule
\multirow{2}{*}{Training Data} & \multirow{2}{*}{Method} & \multicolumn{2}{c}{\underline{Sintel (train)}} &  \multicolumn{2}{c}{\underline{KITTI-15 (train)}} & \multicolumn{2}{c}{\underline{Sintel (test)}} & \multicolumn{1}{c}{\underline{KITTI-15 (test)}} \\
& & Clean & Final & F1-epe & F1-all & Clean & Final & F1-all \\
\midrule    
                       - & FlowFields\cite{flowfields}     & -     & -  & - & -  & 3.75  & 5.81 & 15.31 \\
                       - & FlowFields++\cite{flowfields++}     & - & - & - & -  & 2.94 & 5.49 & 14.82 \\
                       S & DCFlow\cite{dcflow}     & -     & - & - & -   & 3.54  & 5.12 & 14.86 \\
                       S & MRFlow\cite{mrflow}         & -  & -  & - & - & 2.53  & 5.38 & 12.19 \\ \midrule
\multirow{9}{*}{C + T} 
                       & HD3\cite{hd3}            & 3.84  & 8.77 & 13.17 & 24.0 & - & - & - \\
                       & LiteFlowNet\cite{liteflownet}    & 2.48  & 4.04  & 10.39 & 28.5 & - & - & - \\
                       & PWC-Net\cite{pwcnet}        & 2.55  & 3.93 & 10.35 & 33.7 & - & - & - \\
                       & LiteFlowNet2\cite{liteflownet2}   & 2.24  & 3.78  & 8.97 & 25.9 & - & - & - \\
                       & VCN\cite{vcn}            & 2.21  & 3.68  & 8.36 & 25.1 & - & -     & - \\ 
                       & MaskFlowNet\cite{maskflownet} & 2.25 & 3.61 & - & \underline{23.1} & - & - & - \\ 
                       & FlowNet2\cite{ilg2017flownet}       & \underline{2.02}  & \ 3.54$^1$ & 10.08 & 30.0 & 3.96  & 6.02 & - \\
                       & Ours (small)   & 2.21  & \underline{3.35} & \underline{7.51} & 26.9 & - & -     & - \\
                       & Ours (2-view)        & \textbf{1.43} & \textbf{2.71} & \textbf{5.04} & \textbf{17.4} & - & - & - \\ \midrule
\multirow{5}{*}{C+T+S/K} & FlowNet2 \cite{ilg2017flownet}  & (1.45) & (2.01) & (2.30) & (6.8) & 4.16  & 5.74 & 11.48  \\
                     & HD3 \cite{hd3}         & (1.87)     & (1.17) & (1.31) & (4.1)  & 4.79  & 4.67 & 6.55 \\
                     & IRR-PWC \cite{irr}     & (1.92) & (2.51) & (1.63) & (5.3) & 3.84  & 4.58  & 7.65 \\
                     & ScopeFlow\cite{scopeflow} & - & - & - & - & \underline{3.59} & \underline{4.10} & \underline{6.82} \\
                     & Ours (2-view) & (0.77) & (1.20) & (0.64) & (1.5) & \textbf{2.08} & \textbf{3.41} & \textbf{5.27} \\ \midrule
\multirow{6}{*}{C+T+S+K+H}
                     & LiteFlowNet2$^2$ \cite{liteflownet2} & (1.30) & (1.62) & (1.47) & (4.8) & 3.48  & 4.69 & 7.74 \\
                     & PWC-Net+\cite{pwcnet+}   & (1.71)     & (2.34)  & (1.50) & (5.3)  & 3.45  & 4.60 & 7.72 \\
                     & VCN \cite{vcn}            & (1.66)     & (2.24) & (1.16) & (4.1) & 2.81  & 4.40 & 6.30 \\
                     & MaskFlowNet\cite{maskflownet} & - & - & - & - & 2.52 & 4.17 & \underline{6.10} \\
                     & Ours (2-view)       & (0.76)  & (1.22) & (0.63) & (1.5) & \underline{1.94} & \underline{3.18} & \textbf{5.10} \\ 
                     & Ours (warm-start) & (0.77) & (1.27) & - & - & \textbf{1.61} & \textbf{2.86}  & - \\
                     \bottomrule
\end{tabular}
}
\caption{Results on Sintel and KITTI datasets. We test the generalization performance on Sintel(train) after training on FlyingChairs(C) and FlyingThing(T), and outperform all existing methods on both the clean and final pass. The bottom two sections show the performance of our model on public leaderboards after dataset specific finetuning. S/K includes methods which use only Sintel data for finetuning on Sintel and only KITTI data when finetuning on KITTI. +S+K+H includes methods which combine KITTI, HD1K, and Sintel data when finetuning on Sintel. Ours (warm-start) ranks 1st on both the Sintel clean and final passes, and 1st among all flow approaches on KITTI.  ($^1$FlowNet2 originally reported results on the disparity split of Sintel, 3.54 is the EPE when their model is evaluated on the standard data \cite{liteflownet}. $^2$ \cite{liteflownet2} finds that HD1K data does not help significantly during Sintel finetuning and reports results without it. )}
\label{table:Results}
\end{table}

When using C+T for training, our method outperforms all existing approaches, despite using a significantly shorter training schedule. Our method achieves an average EPE (end-point-error) of 1.43 on the Sintel(train) clean pass, which is a 29\% lower error than FlowNet2. These results demonstrates good cross dataset generalization. One of the reasons for better generalization is the structure of our network. By constraining optical flow to be the product of a series of identical update steps, we force the network to learn an update operator which mimics the updates of a first-order descent algorithm. This constrains the search space, reduces the risk of over-fitting, and leads to faster training and better generalization.

When evaluating on the Sintel(test) set, we finetune on the combined clean and final passes of the training set along with KITTI and HD1K data. Our method ranks 1st on both the Sintel clean and final passes, and outperforms all prior work by 0.9 pixels (36\%) on the clean pass and 1.2 pixels (30\%) on the final pass. We evaluate two versions of our model, Ours (two-frame) uses zero initialization, while Ours (warp-start) initializes flow by forward projecting the flow estimate from the previous frame. Since our method operates at a single resolution, we can initialize the flow estimate to utilize motion smoothness from past frames, which cannot be easily done using the coarse-to-fine model.

\subsection{KITTI}
We also evaluate RAFT on KITTI and provide results in Table \ref{table:Results} and Figure \ref{fig:KITTI}. We first evaluate cross-dataset generalization by evaluating on the KITTI-15 (train) split after training on Chairs(C) and FlyingThings(T). Our method outperforms prior works by a large margin, improving EPE (end-point-error) from 8.36 to 5.04, which shows that the underlying structure of our network facilitates generalization. Our method ranks 1st on the KITTI leaderboard among all optical flow methods.

\subsection{Ablations}
\label{sec:ablations}
We perform a set of ablation experiments to show the relative importance of each component. All ablated versions are trained on FlyingChairs(C) + FlyingThings(T). Results of the ablations are shown in Table \ref{table:Ablations}. In each section of the table, we test a specific component of our approach in isolation, the settings which are used in our final model is underlined. Below we describe each of the experiments in more detail.

\setlength\tabcolsep{.7em}
\begin{table}[t]
\centering
\resizebox{0.85\textwidth}{!}{
\begin{tabular}{clccccc}
\toprule
\multirow{2}{*}{Experiment} & \multirow{2}{*}{Method} & \multicolumn{2}{c}{\underline{Sintel (train)}} & \multicolumn{2}{c}{\underline{KITTI-15 (train)}} & \multirow{2}{*}{Parameters} \\
& & Clean & Final & F1-epe & F1-all &  \\
\midrule 
\midrule
\rule{0pt}{1ex} \\
\multicolumn{5}{c}{\emph{Reference Model} (bilinear upsampling), Training: 100k(C) $\rightarrow$ 60k(T)} \\
\midrule
\multirow{2}{*}{Update Op.}  & \underline{ConvGRU} & 1.63 & 2.83 & 5.54 & 19.8 & 4.8M \\
                            & Conv & 2.04 & 3.21 & 7.66 & 26.1 & 4.1M \\ \midrule
\multirow{2}{*}{Tying}      & \underline{Tied Weights} & 1.63 & 2.83 & 5.54 & 19.8 & 4.8M \\
                            & Untied Weights & 1.96 & 3.20 & 7.64 & 24.1 & 32.5M \\ \midrule
\multirow{2}{*}{Context}    & \underline{Context}& 1.63 & 2.83 & 5.54 & 19.8 & 4.8M \\
                            & No Context & 1.93 & 3.06 & 6.25 & 23.1 & 3.3M \\ \midrule
\multirow{2}{*}{Feature Scale}        & \underline{Single-Scale} & 1.63 & 2.83 & 5.54 & 19.8 & 4.8M \\
                                        & Multi-Scale & 2.08 & 3.12 & 6.91 & 23.2 & 6.6M \\ \midrule
\multirow{4}{*}{Lookup Radius}    & 0 & 3.41 & 4.53 & 23.6 & 44.8 & 4.7M \\
                                    & 1 & 1.80 & 2.99 & 6.27 & 21.5 & 4.7M\\
                                    & 2 & 1.78 & 2.82 & 5.84 & 21.1 & 4.8M \\
                                    & \underline{4} & 1.63 & 2.83 & 5.54 & 19.8 & 4.8M \\ \midrule
\multirow{2}{*}{Correlation Pooling}    & No & 1.95 & 3.02 & 6.07 & 23.2 & 4.7M \\
                                        & \underline{Yes} & 1.63 & 2.83 & 5.54 & 19.8 & 4.8M \\ \midrule
\multirow{4}{*}{Correlation Range}      & 32px & 2.91 & 4.48 & 10.4 & 28.8 & 4.8M\\
                                        & 64px & 2.06 & 3.16 & 6.24 & 20.9 & 4.8M\\
                                        & 128px & 1.64 & 2.81 & 6.00 & 19.9 & 4.8M\\
                                        & \underline{All-Pairs} & 1.63 & 2.83 & 5.54 & 19.8 & 4.8M \\  \midrule
\multirow{2}{*}{Features for Refinement}         & \underline{Correlation} & 1.63 & 2.83 & 5.54 & 19.8 & 4.8M \\ 
                                        & Warping & 2.27 & 3.73 & 11.83 & 32.1 & 2.8M \\ \midrule

\rule{0pt}{2ex} \\
\multicolumn{5}{c}{\emph{Reference Model} (convex upsampling), Training: 100k(C) $\rightarrow$ 100k(T)} \\
\midrule
\multirow{2}{*}{Upsampling}         & \underline{Convex} & 1.43 & 2.71 & 5.04  & 17.4 & 5.3M \\
                                        & Bilinear & 1.60 & 2.79 & 5.17 & 19.2 & 4.8M \\ \midrule
\multirow{5}{*}{Inference Updates}  & 1 & 4.04 & 5.45 & 15.30 & 44.5 & 5.3M \\
                                  & 3 & 2.14 & 3.52 & 8.98 & 29.9 & 5.3M \\
                                  & 8 & 1.61 & 2.88 & 5.99 & 19.6 & 5.3M \\
                                  & \underline{32} & 1.43 & 2.71 & 5.00 & 17.4 & 5.3M \\ 
                                  & 100 & 1.41 & 2.72 & 4.95 & 17.4 & 5.3M \\
                                  & 200 & 1.40 & 2.73 & 4.94 & 17.4 & 5.3M \\ \midrule
\end{tabular}
}
\caption{Ablation experiments. Settings used in our final model are underlined. See Sec.~\ref{sec:ablations} for details.}
\label{table:Ablations}
\end{table}

\smallskip \noindent \textbf{Architecture of Update Operator: } We use a gated activation unit based on the GRU cell. We experiment with replacing the convolutional GRU with a set of 3 convolutional layers with ReLU activation. We achieve better performance by using the GRU block, likely because the gated activation makes it easier for the sequence of flow estimates to converge.

\smallskip \noindent \textbf{Weight Tying: } By default, we tied the weights across all instances of the update operator. Here, we test a version of our approach where each update operator learns a separate set of weights. Accuracy is better when weights are tied and the parameter count is significantly lower.

\smallskip \noindent \textbf{Context: } We test the importance of context by training a model with the context network removed. Without context, we still achieve good results, outperforming all existing works on both Sintel and KITTI. But context is helpful. Directly injecting image features into the update operator likely allows spatial information to be better aggregated within motion boundaries.

\smallskip \noindent \textbf{Feature Scale: } By default, we extract features at a single resolution. We also try extracting features at multiple resolutions by building a correlation volume at each scale separately. Single resolution features simplifies the network architecture and allows fine-grained matching even at large displacements. 

\smallskip \noindent \textbf{Lookup Radius: } The lookup radius specifies the dimensions of the grid used in the lookup operation. When a radius of 0 is used, the correlation volume is retrieved at a single point. Surprisingly, we can still get a rough estimate of flow when the radius is 0, which means the network is learning to use 0'th order information. However, we see better results as the radius is increased.

\smallskip \noindent \textbf{Correlation Pooling: } We output features at a single resolution and then perform pooling to generate multiscale volumes. Here we test the impact when this pooling is removed. Results are better with pooling, because large and small displacements are both captured.

\smallskip \noindent \textbf{Correlation Range: } Instead of all-pairs correlation, we also try constructing the correlation volume only for a local neighborhood around each pixel. We try a range of 32 pixels, 64 pixels, and 128 pixels. Overall we get the best results when the all-pairs are used, although a 128px range is sufficient to perform well on Sintel because most displacements fall within this range. That said, all-pairs is still preferable because it eliminates the need to specify a range. It is also more convenient to implement: it can be computed using matrix multiplication allowing our approach to be implemented entirely in PyTorch.  

\smallskip \noindent \textbf{Features for Refinement: } We compute visual similarity by building a correlation volume between all pairs of pixels. In this experiment, we try replacing the correlation volume with a warping layer, which uses the current estimate of optical flow to warp features from $I_2$ onto $I_1$ and then estimates the residual displacement. While warping is still competitive with prior work on Sintel, correlation performs significantly better, especially on KITTI.

\smallskip \noindent \textbf{Upsampling: } RAFT outputs flow fields at 1/8 resolution. We compare bilinear upsampling to our learned upsampling module. The upsampling module produces better results, particularly near motion boundaries.

\smallskip \noindent \textbf{Inference Updates: } Although we unroll 12 updates during training, we can apply an arbitrary number of updates during inference. In Table \ref{table:Ablations} we provide numerical results for selected number of updates, and test an extreme case of 200 to show that our method doesn't diverge. Our method quickly converges, surpassing PWC-Net after 3 updates and FlowNet2 after 6 updates, but continues to improve with more updates.

\subsection{Timing and Parameter Counts}
\begin{figure}[h]
    \centering
	\includegraphics[width=\columnwidth]{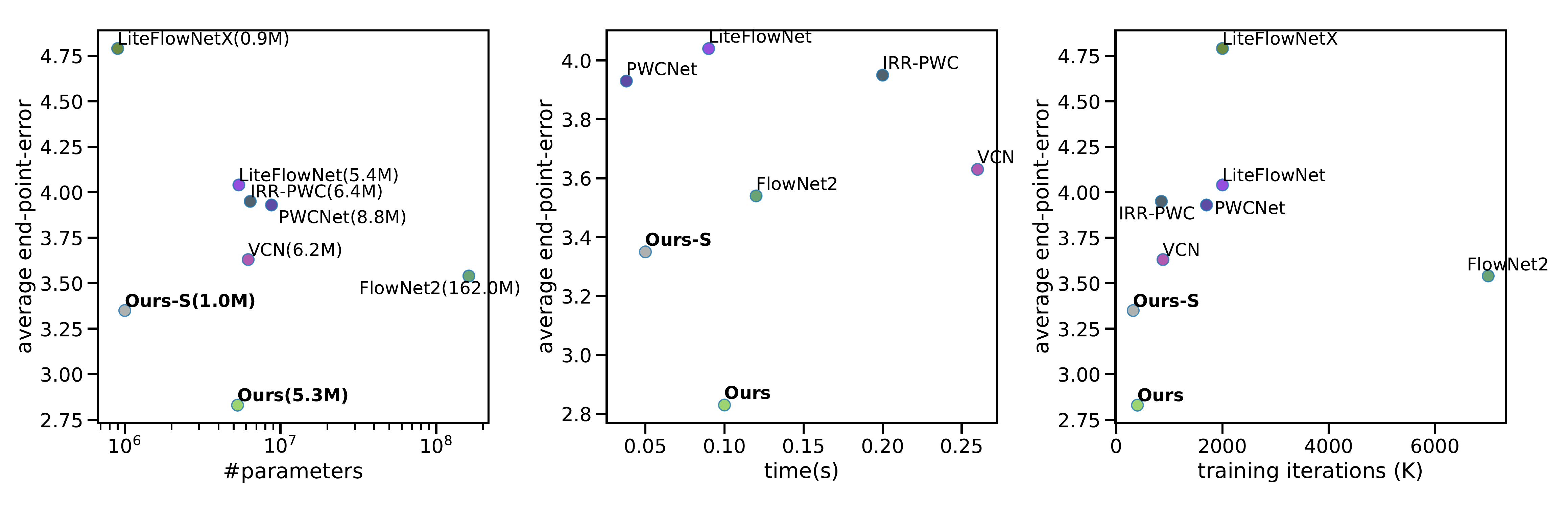}
	\caption{Plots comparing parameter counts, inference time, and training iterations vs. accuracy. Accuracy is measured by the EPE on the Sintel(train) final pass after training on C+T. \emph{Left}: Parameter count vs. accuracy compared to other methods. RAFT is more parameter efficient while achieving lower EPE. \emph{Middle}: Inference time vs. accuracy timed using our hardware \emph{Right}: Training iterations vs. accuracy (taken as product of iterations and GPUs used).}
	\label{fig:ParametersTimingIterations}
\end{figure}

Inference time and parameter counts are shown in Figure \ref{fig:ParametersTimingIterations}. Accuracy is determined by performance on the Sintel(train) final pass after training on FlyingChairs and FlyingThings (C+T). In these plots, we report accuracy and timing after 10 iterations, and we time our method using a GTX 1080Ti GPU. Parameters counts for other methods are taken as reported in their papers, and we report times when run on our hardware. RAFT is more efficient in terms of parameter count, inference time, and training iterations. Ours-S uses only 1M parameters, but outperforms PWC-Net and VCN which are more than 6x larger. We provide an additional table with numerical values for parameters, timing, and training iterations in the supplemental material.

\begin{figure}[h]
    \centering
	\includegraphics[width=\columnwidth]{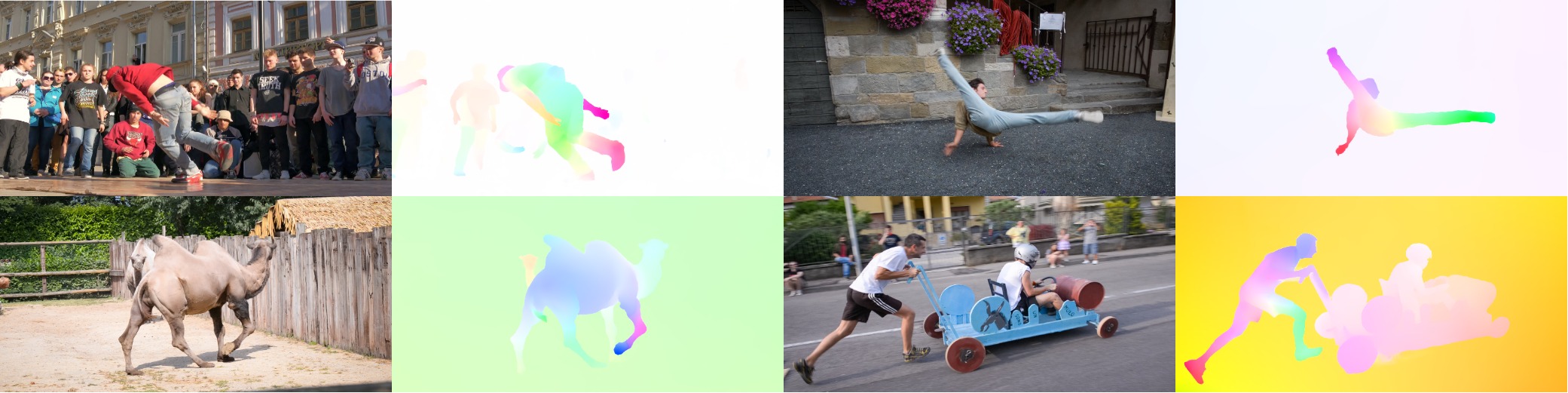}
	\caption{Results on 1080p (1088x1920) video from DAVIS (550 ms per frame).}
	\label{fig:DAVIS}
\end{figure}

\subsection{Video of Very High Resolution}
To demonstrate that our method scales well to videos of very high resolution we apply our network to HD video from the DAVIS\cite{davis} dataset. We use 1080p (1088x1920) resolution video and apply 12 iterations of our approach. Inference takes 550ms for 12 iterations on 1080p video, with all-pairs correlation taking 95ms. Fig.~\ref{fig:DAVIS} visualizes example results on DAVIS. 

\section{Conclusions}
We have proposed RAFT---Recurrent All-Pairs Field Transforms---a new end-to-end trainable model for optical flow.  RAFT is unique in that it operates at a single resolution using a large number of lightweight, recurrent update operators. Our method achieves state-of-the-art accuracy across a diverse range of datasets, strong cross dataset generalization, and is efficient in terms of inference time, parameter count, and training iterations.

\smallskip \noindent \textbf{Acknowledgments: }  This  work was partially funded by the National Science Foundation under Grant No. 1617767.

\bibliographystyle{splncs04}
\bibliography{egbib}

\clearpage

\appendix

\section{Network Architecture}
\vspace{-5mm}

\begin{figure}[h]
    \centering
	\includegraphics[width=.9\columnwidth]{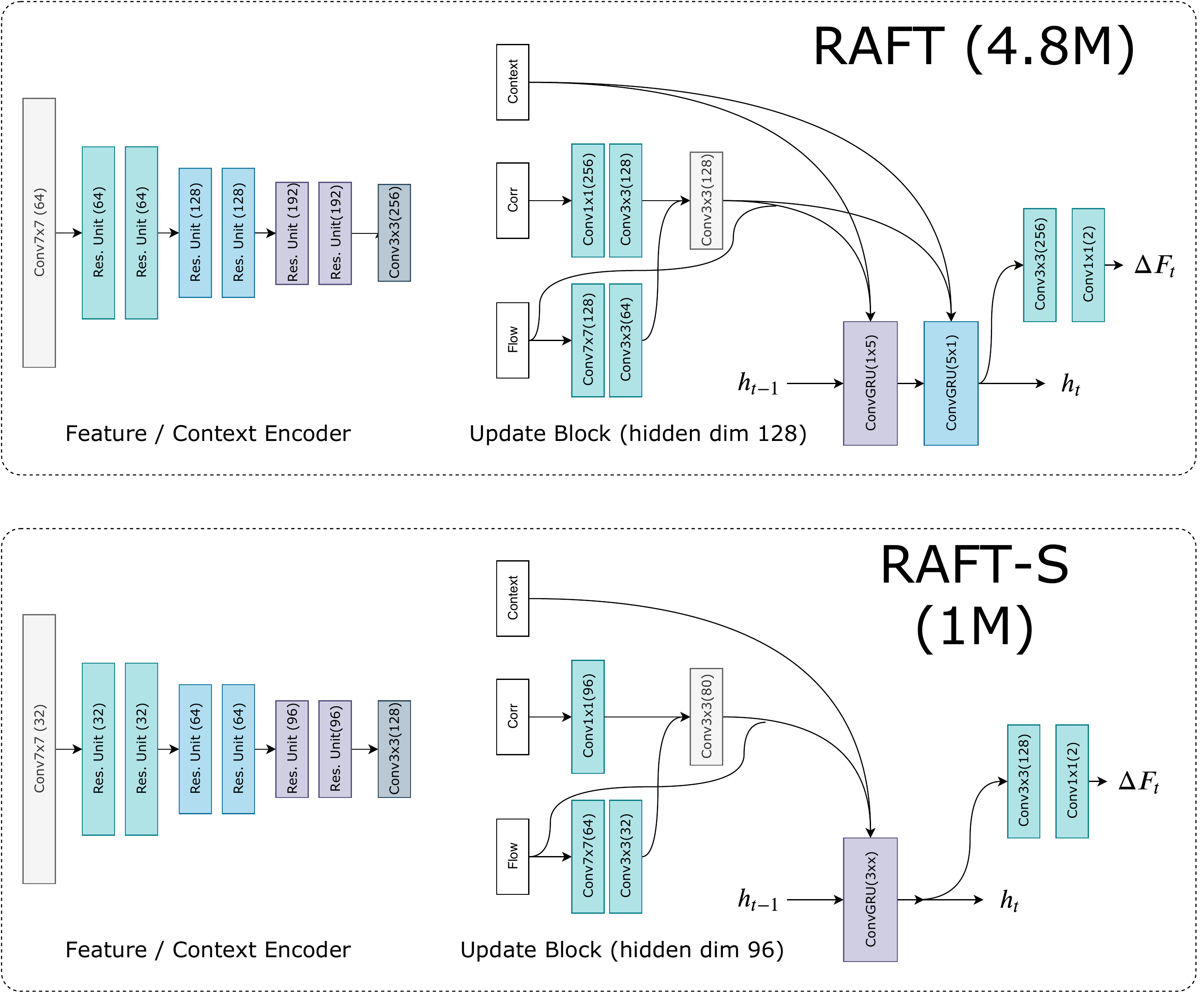}
	\vspace{-3mm}
	\caption{Network architecture details for the full 4.8M parameter model (5.3M with upsampling module) and the small 1.0M parameter model. The context and feature encoders have the same architecture, the only difference is that the feature encoder uses instance normalization while the context encoder uses batch normalization. In RAFT-S, we replace the residual units with bottleneck residual units. The update block takes in context features, correlation features, and flow features to update the latent hidden state. The updated hidden status is used to predict the flow update. The full model uses two convolutional GRU update blocks with 1x5 filters and 5x1 filters respectively, while the small model uses a single GRU with 3x3 filters.}
	\label{fig:A1}
	\vspace{-6mm}
\end{figure}

\clearpage

\section{Upsampling Module}
\vspace{-5mm}

\begin{figure}[h!]
    \centering
	\includegraphics[width=.7\columnwidth]{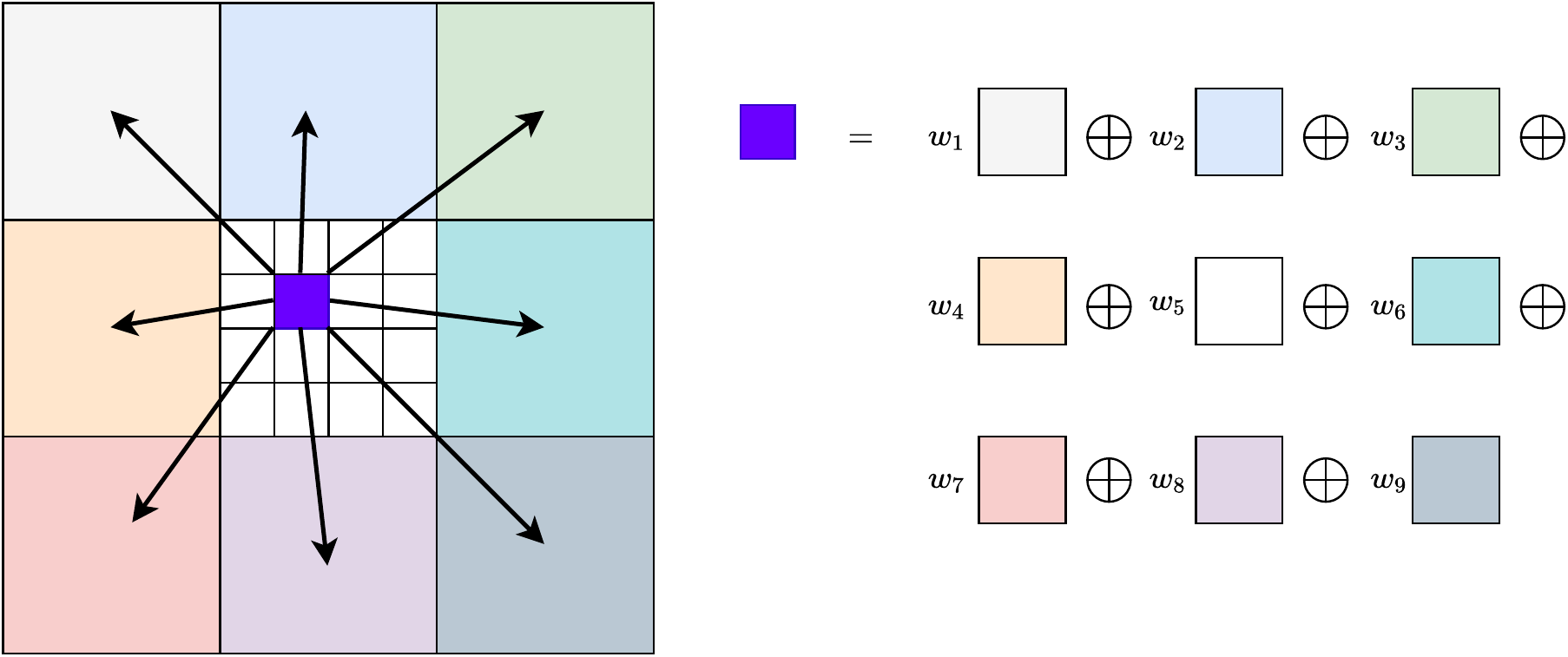}
	\caption{Illistration of the upsampling module. Each pixel of the high resolution flow field (small boxes) is taken to be the convex combination of its 9 coarse resolution neighbors using weights predicted by the network.}
	\label{fig:Convergence}
	\vspace{-6mm}
\end{figure}

\begin{figure}[h!]
	\vspace{-4mm}
	\includegraphics[width=\columnwidth]{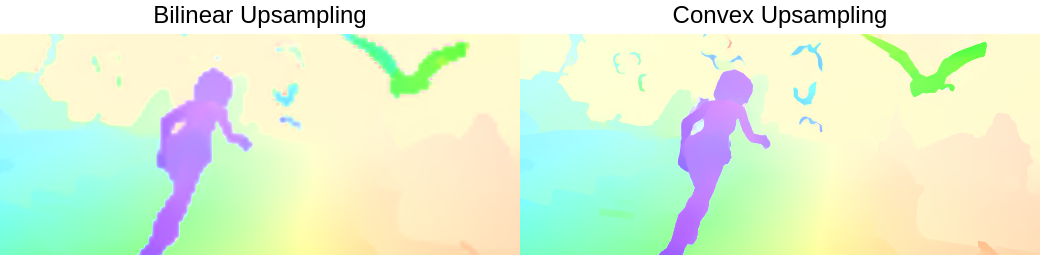}
	\caption{Our upsampling module improves accuracy near motion boundaries, and also allows RAFT to recover the flow of small fast moving objects such as the birds shown in the figure.}
	\label{fig:Upsampling}
	\vspace{-6mm}
\end{figure}

\section{Training Details}
\setlength\tabcolsep{.6em}
\begin{table}[]
\centering
\resizebox{.9\textwidth}{!}{
\begin{tabular}{lcccccc}
\toprule
Stage & Weights & Training Data & Learning Rate & Batch Size (per GPU) &  Weight Decay & Crop Size \\
\midrule
Chairs & - & C & 4e-4 & 6 & 1e-4 & [368, 496] \\
Things & Chairs & T & 1.2e-4 & 3 & 1e-4 & [400, 720] \\
Sintel & Things & S+T+K+H & 1.2e-4 & 3 & 1e-5 & [368, 768] \\
KITTI & Sintel & K & 1e-4 & 3 & 1e-5 & [288, 960] \\

\end{tabular}
}
\vspace{2mm}
\caption{Details of the training schedule. Dataset abbreviations: C: FlyingChairs, T: FlyingThings, S: Sintel, K: KITTI-2015, H: HD1K. During the Sintel Finetuning phase, the dataset distribution is S(.71), T(.135), K(.135), H(.02).}
\label{table:ParamsAndTime}
\vspace{-3mm}
\end{table}

\noindent \textbf{Photometric Augmentation: } We perform photometric augmentation by randomly perturbing brightness, contrast, saturation, and hue. We use the Torchvision \texttt{ColorJitter} with brightness 0.4, contrast 0.4, saturation 0.4, and hue 0.5/$\pi$. On KITTI, we reduce the degree of augmentation to brightness 0.3, contrast 0.3, saturation 0.3, and hue 0.3/$\pi$. With probablity 0.2, color augmentation is performed to each of the images independently. 

\vspace{1mm}
\noindent \textbf{Spatial Augmentation: } We perform spatial augmentation by randomly rescaling and stretching the images. The degree of random scaling depends on the dataset. For FlyingChairs, we perform spatial augmentation in the range $2^{[-0.2, 1.0]}$, FlyingThings $2^{[-0.4, 0.8]}$, Sintel $2^{[-0.2, 0.6]}$, and KITTI $2^{[-0.2, 0.4]}$. Spatial augmentation is performed with probability 0.8. 

\vspace{1mm}
\noindent \textbf{Occlusion Augmentation: } Following HSM-Net \cite{hsm}, we also randomly erase rectangular regions in $I_2$ with probability 0.5 to simulate occlusions.

\section{Timing, Parameters, and Training Iterations}

\begin{figure}[h]
    \vspace{-8mm}
	\hspace{-2mm}\includegraphics[width=0.45\columnwidth]{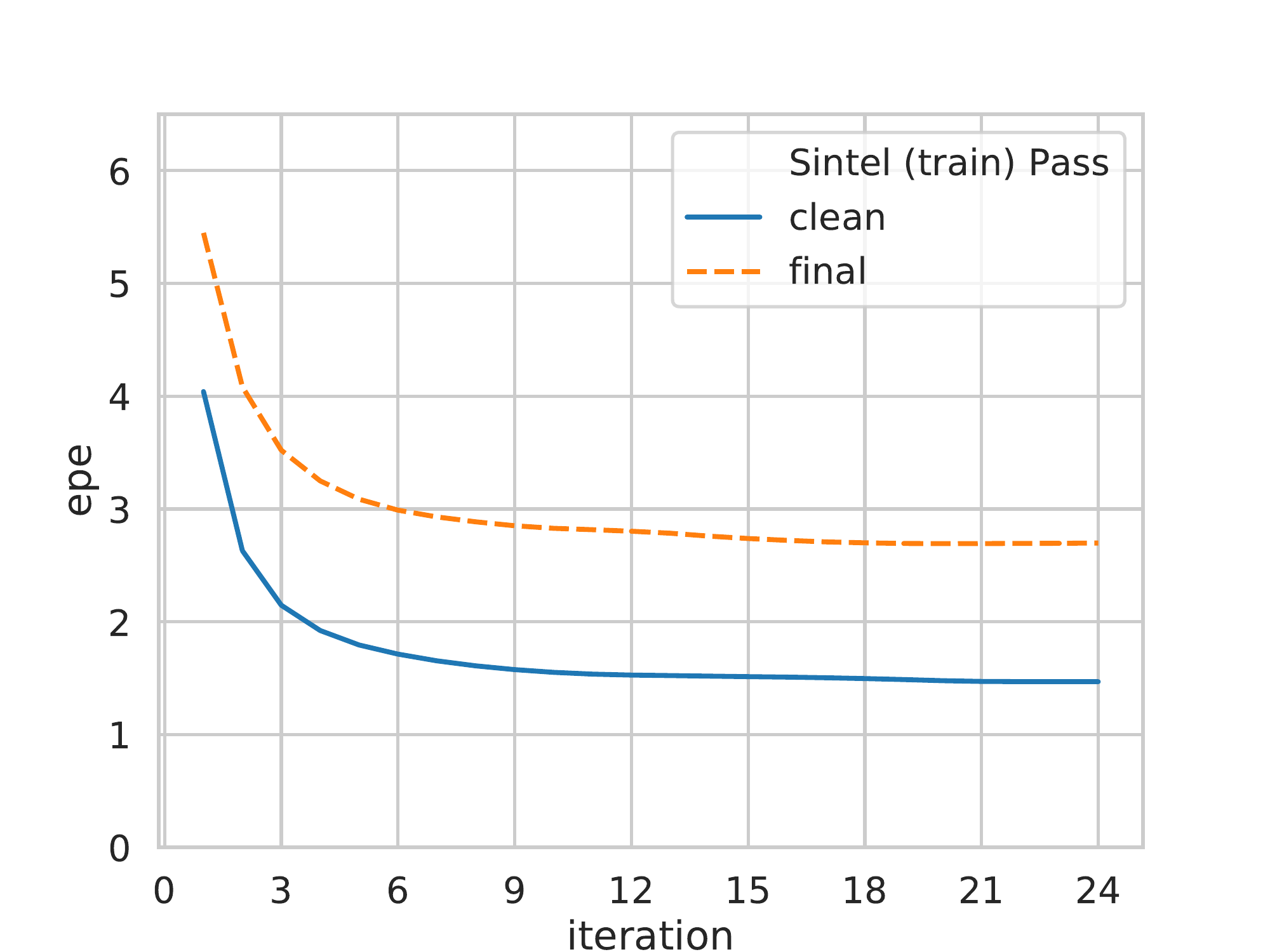}
	\hspace{-2mm}\includegraphics[width=0.57\columnwidth]{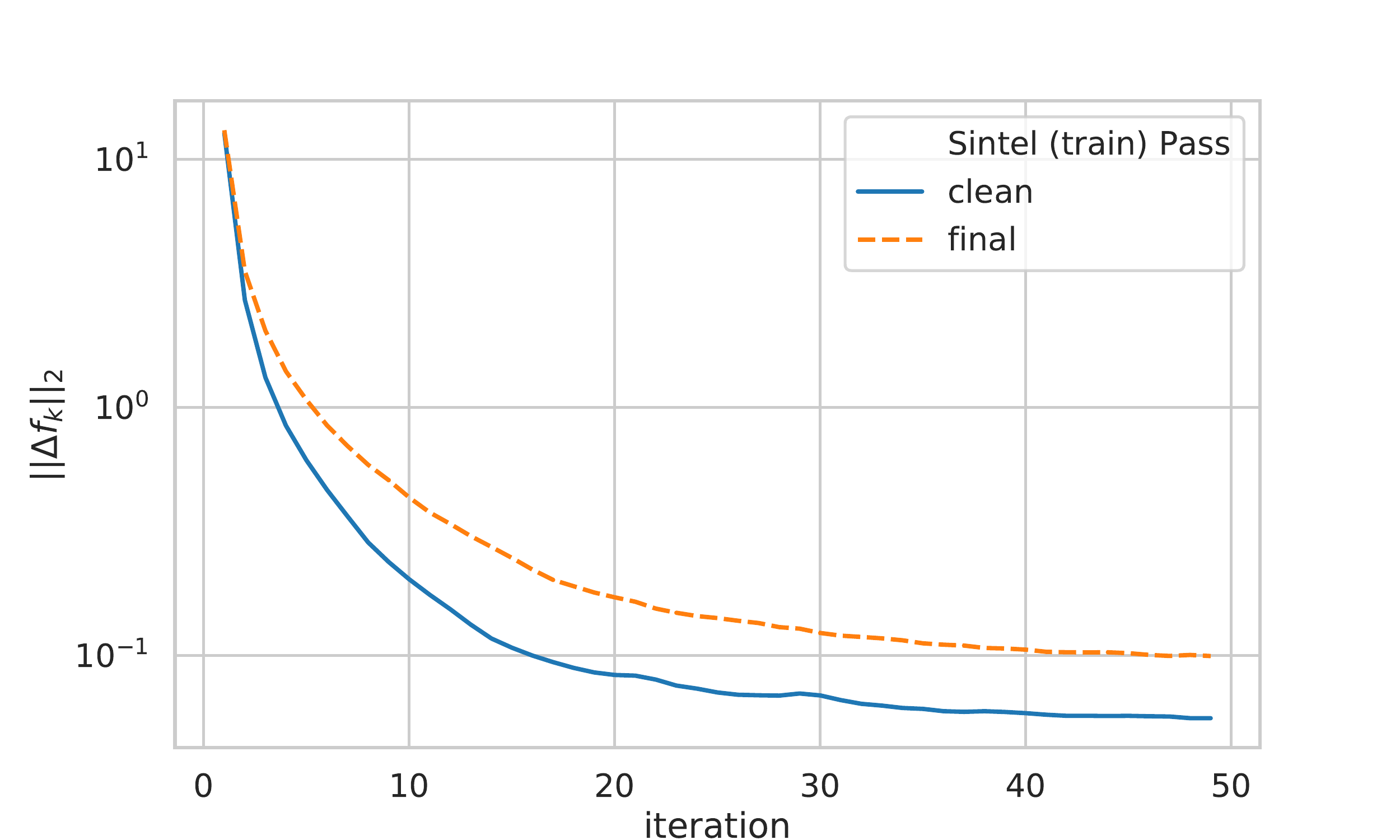}
	\vspace{-3mm}
	\caption{(Left) EPE on the Sintel set as a function of the number of iterations at inference time. (Right) Magnitude of each update $||\Delta \mathbf{f}_k||_2$ averaged over all pixels indicating convergence to a fixed point $\mathbf{f}_k \rightarrow \mathbf{f}^*$.}
	\label{fig:Convergence}
	\vspace{-6mm}
\end{figure}

\setlength\tabcolsep{.6em}
\begin{table}[]
\centering
\resizebox{.9\textwidth}{!}{
\begin{tabular}{lccccc}
\toprule
Method & Parameters (M) & Time (Reported) & Time (1080Ti) & Training Iter. (\#GPUs)& Accuracy \\ \midrule
LiteFlowNetX\cite{liteflownet} & 0.9M & 0.03s & - & 2000k & 4.79 \\
LiteFlowNet\cite{liteflownet} & 5.4M & 0.09s & 0.09s & 2000k & 4.04 \\
IRR-PWC\cite{irr} & 6.4M & - & 0.20s & 850k & 3.95 \\
PWCNet+\cite{pwcnet+} & 9.4M & 0.03s & 0.04s & 1700k &  3.93 \\
VCN\cite{vcn} & 6.2M & 0.18s & 0.26s & 220k(4) & 3.63 \\ 
FlowNet2\cite{ilg2017flownet} & 162M & 0.12s & 0.11s & 7000k & 3.54 \\ \midrule
Ours (small) & 1.0M & - & 0.05s & 160k(2) & 3.37 \\
Ours (mixed) & 5.3M & - &  0.10s & 240k(1) & 2.85 \\
Ours & 5.3M & - & 0.10s & 200k(2) & 2.83 \\

\end{tabular}
}
\caption{Parameter counts, inference time, training iterations, and accuracy on the Sintel (train) final pass. We report the timing and accuracy of our method after 10 updates using a GTX 1080Ti GPU. If possible, we download the code from the other methods and re-time using our machine. If the model is trained using more than one GPU, we report the number of GPUs used to train in parenthesis. We can also train RAFT using mixed precision training Ours(mixed) and achieve similar results while training on only a single GPU. Overall, RAFT requires fewer training iterations and parameters when compared to prior work. }
\label{table:ParamsAndTime}
\vspace{-3mm}
\end{table}
\clearpage

\end{document}